%% file: main.tex
\documentclass[runningheads]{llncs}

\usepackage{eccv}

\usepackage{eccvabbrv}

\usepackage{graphicx}
\usepackage{booktabs}

\usepackage{algorithm}
\usepackage{algpseudocode}
\usepackage{arydshln}

\algrenewcommand{\algorithmiccomment}[1]{\hfill{\footnotesize{\textcolor{teal!70!black}{\(\triangleright\) #1}}}}

\usepackage{hyperref}

\usepackage{orcidlink}
\newcommand{\netnarrow}{narrow\xspace}
\newcommand{\netNarrow}{Narrow\xspace}

\newcommand{\ours}{RBDC\xspace}

\newcommand{\netnarrower}{narrower\xspace}
\newcommand{\netnarrowest}{narrowest\xspace}

\newcommand{\netwide}{wide\xspace}
\newcommand{\netWide}{Wide\xspace}

\newcommand{\netwider}{wider\xspace}

\newcommand{\flop}{FLOP\xspace}
\newcommand{\flops}{FLOPs\xspace}
\newcommand{\gflops}{GFLOPs\xspace}

\newcommand{\first}{1\textsuperscript{st}\xspace}
\newcommand{\second}{2\textsuperscript{nd}\xspace}

\newcommand{\fourth}{4\textsuperscript{th}\xspace}

\usepackage{titletoc}

\begin{document}

\newcommand{\papertitle}{Recursive Block-Diagonal Coupling\\for Resource-Efficient Training of Vision Models}
\title{\papertitle}

\titlerunning{Resource-Efficient Training of Vision Models}

\author{Maxim Henry\orcidlink{0009-0007-8899-8723} \and
Adrien Deli\`ege\orcidlink{0000-0003-3981-6982} \and
S\'ebastien Pi\'erard\orcidlink{0000-0001-8076-1157} \and\\
Marc Van Droogenbroeck\orcidlink{0000-0001-6260-6487}}

\authorrunning{M.~Henry et al.}

\institute{Montefiore Institute, University of Li\`ege, Li\`ege, Belgium\\
\{Maxim.Henry, Adrien.Deliege, S.Pierard, M.VanDroogenbroeck\}@uliege.be}
\maketitle

\newcommand{\epochsNarrow}{\textrm{epochs}_\textrm{\netnarrow}}
\newcommand{\epochsWide}{\textrm{epochs}_\textrm{\netwide}}

\input{sections/0_abstract}
\input{sections/1_introduction}

\input{sections/2_related_work_alternative}
\input{sections/3_method}

\input{sections/4_results}

\input{sections/5_conclusion}

\makeatletter
\@ifpackagewith{eccv}{review}{
}{
    \subsection*{Acknowledgements}
    M. Henry and S. Pi{\'e}rard are funded by grant 2010235 (ARIAC by \href{https://www.digitalwallonia.be/en/}{DIGITALWALLONIA4.AI}) of the SPW EER, Wallonia, Belgium; A. Deli{\`e}ge is a \href{https://www.frs-fnrs.be}{F.R.S.-FNRS} postdoc researcher.The present research benefited from computational resources made available on \emph{Lucia}, the \emph{Tier-1} supercomputer of the Walloon Region, infrastructure funded by the Walloon Region under the grant agreement n°1910247.
}
\makeatother

\newpage
{
    \small
    \bibliographystyle{splncs04-MVD}
    \bibliography{bib/abbreviation-short,bib/abbrevbiation-empty,
    bib/all,
    bib/PUTS_YOUR_NEW_REFS_HERE}
}

\input{sections/A_0_supplementary_material}

\end{document}

%% file: sections/0_abstract.tex
\begin{abstract}

Training high-capacity vision models from scratch requires substantial computational resources. %
To improve training efficiency of a \netwide target model, existing growth methods often assume the availability of \netnarrower models, obscuring the true computational cost of the entire pipeline. 
We propose an efficient training protocol, \ours, that builds \netwide models by coupling in a parameter-free block-diagonal way \netnarrower, independently trained models in a recursive way. %
This allows a flexible allocation of the training budget available across all the models involved. %
Evaluated with vision transformers (DeiT) and convolutional networks (ResNet) on ImageNet, our \ours training protocol shows a much better efficiency than models trained from scratch with the standard protocol, yielding 30\% \flops reduction at similar test accuracies. It also achieves higher performances at same training \flops than training protocols from the model growth literature. 
Finally, we show that our models can serve as better backbones than their original counterparts for downstream object detection and instance segmentation tasks. %

\keywords{Efficient Training \and Recursive Training \and \flops{} \and Transformer \and CNN \and Model Weights Initialization}
\end{abstract}

\begin{figure*}[t!]
    \centering
    \includegraphics[width=0.80\linewidth]{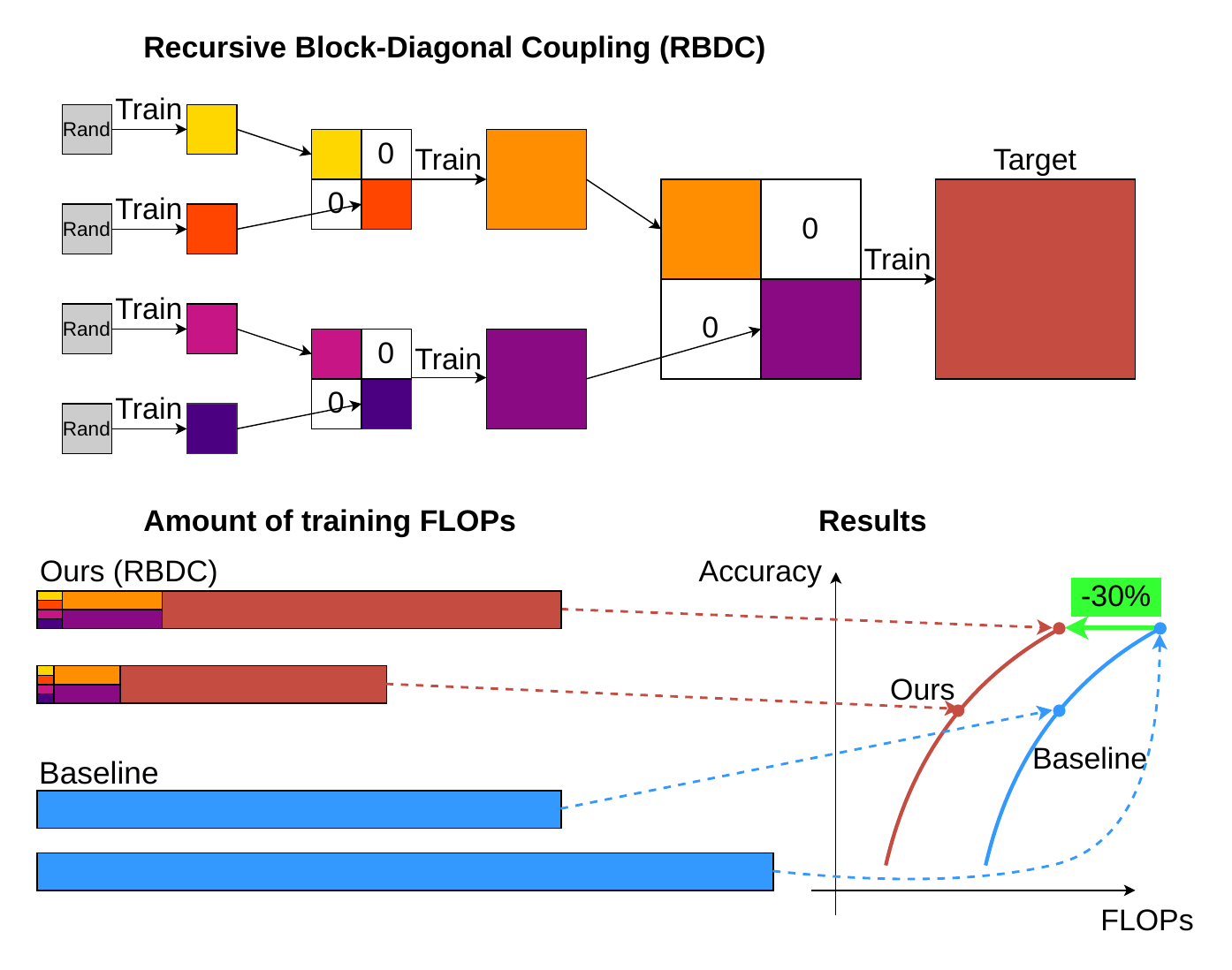}
    \caption{
        \textbf{Overview.} We introduce a resource-efficient training protocol for high-capacity computer vision models (\eg, transformers, CNNs) that we call \emph{Recursive Block-Diagonal Coupling} (\ours). To train a target model, \ours recursively trains \netnarrower models and couples them to build a block-diagonal initialization for training \netwider models (top). %
        The budget of training \flops is distributed according to the different model sizes (bottom left). Our results show that \ours is much more resource-efficient than  the standard randomly initialized training protocol (bottom right). %
    }
    \label{fig:graphabstract}
\end{figure*}

%% file: sections/1_introduction.tex
\section{Introduction}
\label{sec:intro}

Deep learning has achieved remarkable success across computer vision tasks, largely driven by the upscaling of models. However, model upscaling requires a proportional increase in computing power. Hardware limits are approached, and the financial cost of training high-capacity models from scratch and using them grows~\cite{Thompson2024AModel}. It is thus essential to improve the efficiency of such models, and in our case, those used in computer vision (\eg, transformers, CNNs).

According to the Sustainability Directory~\cite{Sustainability2026Resource}, ``resource-efficient machine learning, at its core, is about optimizing machine learning processes to minimize resource consumption without sacrificing performance''. However, a distinction must be made between the training efficiency and the inference efficiency. In this work, we aim at improving resource efficiency during training in the sense of improving the trade-off curve between a model's performance and the computational resources consumed during training. Specifically, we measure performance via test accuracy and resources via floating-point operations (\flops).

Training efficiency depends on several elements that govern the overall training protocol (\eg, number of epochs, model architecture, dataset size, etc.), and various protocols can be defined. 
For example, the \emph{standard protocol} to train a vision model from scratch consists of randomly initializing its weights and then training them. We call the final, trained model ready for deployment the \emph{target model}. The standard protocol is not the only way to obtain it.
Wide-to-narrow approaches, such as pruning, initially train a \emph{\netwide model} (with far more parameters than the target model) and prune it into the \netnarrower target model. Conversely, narrow-to-wide approaches, such as model growing, begin with a \emph{\netnarrow model} progressively expanded into the \netwider target model. In both cases, the target model remains the final objective, while \emph{\netnarrow} and \emph{\netwide} are relative size descriptors depending on the optimization direction (wide-to-narrow versus narrow-to-wide). Our work is closer to the latter case and, unlike training protocols based on fine-tuning pre-trained models, such as parameter-efficient fine-tuning \cite{Xu2026ParameterEfficient,Houlsby2019Parameter,Hu2022LoRA}, focuses on training models \emph{from scratch}.

\textbf{Contributions.} Our goal is to develop a training protocol that improves the trade-off between the number of \flops used for training and the test accuracy reached by the trained model. In particular, our contributions are threefold.%

First, we introduce \emph{Recursive Block-Diagonal Coupling} (\ours), an efficient training protocol that constructs and trains high-capacity models recursively: distinct \netnarrow models are independently trained and coupled into a \netwide model via block-diagonal initialization with zero-padded interaction terms (see \cref{fig:graphabstract}). This parameter-free initialization %
lets optimization of the \netwide model start from an ``educated'' state, rather than the random initialization used in the standard protocol.%
    
Second, we show that our \ours training protocol is much more efficient than the standard one, as also depicted in \cref{fig:graphabstract}, reducing by up to 30\% the training \flops needed to reach a high test accuracy %
 on ImageNet with major computer vision model architectures, \ie transformers (DeiT) and CNNs (ResNet). %

Third, we show that the benefits of our \ours training protocol extend beyond improving training efficiency. Specifically, models trained with \ours can serve as well-performing pretrained backbones for downstream tasks such as object detection and instance segmentation on the COCO benchmark~\cite{lin2014microsoft}, with performances on par or better than models trained with the standard protocol. %

%% file: sections/2_related_work_alternative.tex
\section{Related Work}
\label{sec:relatedwork}

\subsection{Inference Efficiency vs. Training Efficiency}
\noindent \textbf{Inference Efficiency.} A significant portion of research focuses on the efficiency of a model after deployment (inference)~\cite{Sze2017Efficient, Menghani2023Efficient}, with the objective of maximizing performance within a fixed computational inference budget. Common inference-efficient approaches include structured pruning~\cite{Han2015Learning, Henry2026LinDeps} and knowledge distillation~\cite{Gou2021Knowledge} to reduce model size for deployment. However, such top-down paradigms require an initial (very) \netwide model or teacher, whose training consumes a high amount of resources. Similarly, post-training quantization~\cite{Banner2018PostT4} reduces the precision of model weights to lower memory footprint and latency but remains an inference-centric optimization. While effective for deployment, these techniques generally do not alleviate the computational burden of the initial training phase.

\noindent \textbf{Training Efficiency.} A parallel stream of research, including ours, focuses on methods to improve \emph{training efficiency}~\cite{Gusak2022Survey, Shen2024OnEfficient}. This focus is important because relying on model scaling for performance gains requires an exponential increase in computing power \cite{Thompson2020TheComputational-arxiv}. Furthermore, recent economic models for computer vision systems demonstrate that the financial cost of training \netwide models from scratch can create significant barriers to deployment \cite{Thompson2024AModel}. This computational scaling is evident in standard architectures: progressing from a vision transformer base (ViT-B, 86M parameters, 17.5 \gflops) to a large (307M parameters, 61.5 \gflops) or huge (632M parameters, 167 \gflops) increases training resource consumption~\cite{Touvron2022DeiT}. Efforts to mitigate these costs include parameter-centric and model-centric approaches, as described hereafter.

\subsection{Parameter-Centric Training Efficiency}

\textbf{Low-Precision Training.} 
Distinct from quantization for inference, low-pre\-ci\-sion training aims to accelerate gradient calculation during the optimization. Standard techniques now leverage mixed-precision (FP16/BF16) \cite{Micikevicius2018Mixed}. More aggressive approaches, such as FP8 training \cite{Micikevicius2022FP8Formats-arxiv}, exploit modern hardware to further reduce memory footprint, though they often require specialized stabilization to prevent numerical underflow during backpropagation. Though effective, low-precision training is hardware-dependent and aims to reduce the cost per \flop, while we focus on reducing the amount of \flops. The two can be naturally paired: our \ours training protocol with \netnarrow models and the \netwide target model could in principle be trained in mixed-precision for compounded acceleration.

\noindent \textbf{Parameter-Efficient Fine-Tuning (PEFT)}~\cite{Xu2026ParameterEfficient}\textbf{.} 
Rather than updating all weights during training steps, PEFT methods target a small subset of parameters. Techniques like Adapters \cite{Houlsby2019Parameter} and LoRA  \cite{Hu2022LoRA} freeze the pre-trained backbone and inject small, trainable modules. While these methods reduce the parameters updated per step, they are traditionally reserved for fine-tuning rather than for training from scratch. The block-diagonal coupling of our \ours training protocol draws conceptual inspiration from the zero-initialization of interaction terms in LoRA but applies it much more extensively in a recursive way.

\subsection{Model-Centric Training Efficiency}

\textbf{Progressive Growing.}
Training \netwide models from scratch prevents leveraging \netnarrower models trained with fewer resources. Hence, recent research explores initializing \netwide architectures from pre-trained \netnarrower weights. A common approach is \emph{function-preserving growth}: techniques such as Net2Net \cite{Chen2015Net2Net-arxiv} and progressive stacking \cite{Gong2019Efficient} expand a single model's width or depth while preserving its exact function at expansion. Net2Net, for instance, does this by duplicating existing units and breaking symmetry with small noise perturbations. Other methods, such as GradMax \cite{Evci2022GradMax}, initialize new capacity by zeroing incoming weights to preserve function, while using gradient information to optimize the outgoing weights, letting the \netwider model bypass the initial chaotic phase of convergence. Recent works also apply progressive growth to vision transformers; for example, LiGO~\cite{Li2026Efficient} grows smaller pre-trained weights into larger target networks, while AutoProg~\cite{Wang2023LearningToGrow} combines layer and token growth. 
Additionally, Xia et al.~\cite{xia2023budgeted} propose a budgeted training framework that dynamically expands the network's active capacity—such as the number of attention heads and hidden dimensions—to accommodate a target computational budget. However, these methods typically expand a single model instance, so the expanded model is constrained by its precursor's singular feature space. In contrast, our \ours protocol benefits from the learned knowledge of multiple \netnarrower models.

\textbf{Recombining Learned Parameters.} 
A more recent paradigm involves synthesizing a \netwider model by aggregating multiple \netnarrower, pre-trained models. Unlike standard ensembling, which averages outputs, these methods integrate parameters into a single unified architecture. Approaches like MixtureGrowth~\cite{Pham2024MixtureGrowth} demonstrate that recombining distinct \netnarrower models can act as a powerful initialization for a \netwider student. However, such methods typically assume that at least one of the \netnarrow models is already available in an existing model zoo, using this assumption to justify the training speedup. Accounting for the resources needed to train the \netnarrow models is a saner practice and reveals more moderate benefits. %
In that setting, we show that our \ours training protocol provides an efficiency gain even when accounting for the total computational budget required to train all \netnarrow models completely from scratch.

%% file: sections/3_method.tex
\section{Method}
\label{sec:method}

In line with the model-centric approaches for training efficiency, we propose a recursive training protocol, \ours, that constructs high-capacity target models by combining \netnarrower, independently trained \netnarrow models. Unlike prior recombination protocols that assume the availability of pre-trained models in a zoo~\cite{Chen2015Net2Net-arxiv, Pham2024MixtureGrowth}, our framework is designed to yield  computational efficiency when training entirely from scratch.
Essentially, \ours leverages the observation that training multiple \netnarrower models is computationally cheaper than training a single \netwide model from scratch. By training these \netnarrow models with distinct initializations, we capture a diverse set of features. By initializing the \netwider target model via a block-diagonal formulation to benefit from the independently learned features of the \netnarrow models, we observe that we can afford to effectively reduce the total \flops required to train the model at no performance drop, as we will show in \cref{sec:results}. The pseudocode of \ours, which complements~\cref{fig:graphabstract}, is given in \cref{algo:ours} and is further explained hereafter.

\begin{algorithm}
\caption{Pseudocode of our Recursive Block-Diagonal Coupling (\ours), exemplified in the case of training linear layers.\label{algo:ours}}
\begin{algorithmic}[1]

\State \textbf{const} $r$ \Comment{Training ratio of \netwide and \netnarrow models}

\Function{RBDC}{$\text{size}, \text{epochs}, \text{min\_size}$} \Comment{Our \ours definition}
    \If{$\text{size} < \text{min\_size}$} \Comment{Model is \netnarrow enough, stop recursion}
        \State $M \gets \text{rand}([\text{size}, \text{size}])$ \Comment{Random initialization of \netnarrow models}
        \State $M \gets \text{train}(M, \text{epochs})$ \Comment{Regular model training}
    \Else \Comment{Model is still too \netwide, apply recursion}

        \State $\epochsNarrow \gets \text{epochs}/(r + 2)$  \Comment{Epochs to train \netnarrow models}
        \State $\epochsWide \gets \text{epochs} \cdot r/(r + 2)$ \Comment{Epochs to train \netwide model}

        \State $M_1 \gets \text{RBDC}(\text{size}/2, \epochsNarrow, \text{min\_size})$ \Comment{Train a \netnarrow model}
        \State $M_2 \gets \text{RBDC}(\text{size}/2, \epochsNarrow, \text{min\_size})$ \Comment{Train a \netnarrow model}

        \State $M \gets 
        \begin{bmatrix}
        M_1 & 0 \\
        0 & M_2
        \end{bmatrix}$ \Comment{Block-diagonal coupling initializes \netwide model}

        \State $M \gets \text{train}(M, \epochsWide)$ \Comment{Train \netwide model}
    \EndIf
    \State \Return $M$ \Comment{Return the trained target \netwide model}
\EndFunction

\end{algorithmic}
\end{algorithm}

\subsection{Recursive Width Halving}

\ours training protocol operates on the principle of recursive width halving. Let us consider a target model with width $W$ (\eg, channel count in CNNs or embedding dimension in transformers). Instead of initializing it randomly to train it, we initialize it with two \netnarrower models, each with width $W/2$. These \netnarrow models are themselves initialized with \netnarrower models, which are also initialized with even \netnarrower models, recursively. When the \netnarrowest initializing models are small enough, or cannot be halved anymore, they are randomly initialized with distinct seeds and trained independently following the standard protocol. After training, they are distinct experts capturing potentially different features, and they are then coupled to serve as an educated initialization for training the \netwider model from which they originate.

\subsection{Block-Diagonal Initialization with Zero-Padding}
\label{subsec:initialization}

Our \ours training protocol is applicable to models composed of different types of layers. 
The code used for our experiments includes specific implementations for seven types of layers: attention qkv projection layers, attention output projection layers, position-wise feed-forward layers, head layers, convolution layers, layer normalization layers, and batch normalization layers. 
These types cover most of the modern models used in computer vision (\eg, ViTs and CNNs). Given that most \flops are consumed in linear layers (the first four types of layers listed) and convolution layers, we will focus below on a detailed explanation of \ours for these layers. 
The pseudocode provided in~\cref{algo:ours} and the depiction of \ours in~\cref{fig:graphabstract} apply to these layers, but only the $11^{th}$ line of~\cref{algo:ours} is specific to these cases. Explanations for the other types can be found in supplementary material. 

For coupling two \netnarrow models into a \netwide model, we propose a \emph{block-diagonal initialization} with zero-initialized interaction terms. %
Given a linear layer in the \netnarrow models defined by weight matrices $\mathbf{W}^{(1)}_{\netnarrow}, \mathbf{W}^{(2)}_{\netnarrow} \in \mathbb{R}^{d_{out}/2 \times d_{in}/2}$, we construct the corresponding weight matrix $\mathbf{W}_{\netwide} \in \mathbb{R}^{d_{out} \times d_{in}}$ of the \netwide model as:
\begin{equation}
    \mathbf{W}_{\netwide} = \begin{bmatrix} 
    \mathbf{W}^{(1)}_{\netnarrow} & \mathbf{0} \\
    \mathbf{0} & \mathbf{W}^{(2)}_{\netnarrow} 
    \end{bmatrix}\,, 
    \label{eq:block-diagonal}
\end{equation}
where the weights of the off-diagonal blocks are initialized to zero (and are of course trainable weights). For convolution layers, the block-diagonal initialization is applied across the output and input channel dimensions. For a weight tensor of shape ($C_{out}$, $C_{in}$, $K$,$K$), the learned spatial kernel dimensions ($K\times K$) remain intact, while the cross-model channel connections are zero-padded.

The use of zero initialization for off-diagonal blocks follows other works, notably Net2Net \cite{Chen2015Net2Net-arxiv}, which aims to grow a model by ensuring that identity is preserved before training the grown model. In our case, it indeed ensures that at the start of the \netwide model training phase, the model behaves as an ensemble of the two \netnarrow models. The forward pass of the \netwide model is almost equivalent to the independent execution of the \netnarrow models, up to minor differences in normalization layers' statistics. The zero initialization also echoes parameter-efficient fine-tuning methods like LoRA~\cite{Hu2022LoRA}, which initialize trainable adapters to zero to preserve the pre-trained model's function. In our case, although the entire combined matrix remains fully trainable, this initialization allows the \netwide model to gradually learn the correlations between the features of the \netnarrow models without degrading the initial performance inherited from them. %

\subsection{Efficiency and Training Ratio of \netWide and \netNarrow Models}
\label{subsec:schedule}

We identify that the allocation of the computational budget to train the \netnarrow models and the \netwide model is a parameter mostly overlooked in the literature related to growing models~\cite{Pham2024MixtureGrowth, Evci2022GradMax, Chen2015Net2Net-arxiv}. To investigate training efficiency based on this allocation, we define the following training ratio: 
\begin{equation}
r = \frac{\epochsWide}{\epochsNarrow}\,,
\label{eq:trainingRatio}
\end{equation}
where $\epochsWide$ and $\epochsNarrow$ represent, respectively, the number of epochs used to train the \netwide model and each \netnarrow model. %
As a default value, we recommend using a ratio of $r=2$, meaning that a \netwide model is trained with twice as many epochs as its initializing \netnarrow models. This default choice echoes the width-halving process and is also guided by our experiments, as shown in \ref{subsec:expe-validation-design-choices}. The efficiency in terms of \flops of our \ours training protocol can be tied to the choice of $r$ and to the total budget of epochs available for the whole training. More details on that matter are provided in supplementary material.

%% file: sections/4_results.tex
\section{Results}
\label{sec:results}

\textbf{Target Models.}
We evaluate \ours on both vision transformers (DeiT) and CNNs (ResNet). Throughout this section, we use \emph{tiny}, \emph{small}, \emph{base} to refer to the different sizes of DeiT~\cite{Touvron2021Training, Touvron2022DeiT} and 0.125x, 0.25x, 0.5x, and 1.0x to refer to the different sizes of ResNet-50D~\cite{Zagoruyko2016Wide} where the multiplicator used applies to the width of each convolutional layer, with 1.0x being the target model.

\textbf{Training Efficiency through Normalized \flops.} 
To ensure a rigorous and fair comparison between training protocols, we account for the computational cost of training every model involved, including all \netnarrow models. We use the \texttt{fvcore} library to compute the forward pass cost in multiply-accumulate operations (MACs). 
Because \texttt{fvcore} reports MACs under the \flops nomenclature, we adhere to the definition where 1 MAC represents exactly 2 \flops. 
These values represent the cost of a single forward pass. To estimate the total computational load required for training, we follow \cite{Hoffmann2022Training}: the backward pass requires approximately twice the compute of the forward pass to calculate gradients regarding both activations and weights. 
Therefore, the total compute per training step (one forward and one backward) is three times the forward cost. The total \flops for a complete training phase are thus 3$\times$forward \flops$\times$number of epochs$\times$training images. Throughout this section, we report these costs as \emph{normalized \flops} as done in \cite{Pham2024MixtureGrowth}. 
Specifically, we calculate the training \flops for a given experiment and divide it by the total \flops required to train the baseline target model from scratch (300 epochs for DeiT and 90 epochs for ResNet) with the standard training protocol, providing a clear relative measure of training efficiency.

\textbf{Experiments.} We present our results across $4$ experiments:
(1)~we show the training efficiency of our \ours training protocol in \cref{subsec:expe-efficiency};
(2)~we analyze the impact of training ratio $r$ in \cref{subsec:expe-validation-design-choices};
(3)~we compare \ours against training protocols from the model growth literature on both CNNs and vision transformers in \cref{subsec:expe-comparison-growth}; 
(4)~we validate the usefulness of the models trained with \ours on downstream tasks in \cref{subsec:expe-transfer-learning}.

\subsection{Resource Efficiency of \ours}
\label{subsec:expe-efficiency}

The objective of our \first experiment is to show that, regarding the standard protocol, \ours can train a model with a similar accuracy while requiring a much smaller amount of \flops.

\textbf{Experimental Setup.} 
We use DeiT models~\cite{Touvron2021Training} trained on Image\-Net-1K with the official DeiT codebase and its optimal setup~\cite{facebookresearch-github-DeiT}, which includes Knowledge Distillation (KD). 
Optimization uses AdamW \cite{Loshchilov2019Decoupled} with a base learning rate of 5e-4 and a weight decay of 0.05, and an effective batch size of 1024. 
The learning rate follows a cosine annealing schedule~\cite{loshchilov2017sgdr}, preceded by a 5-epoch linear warmup. For \ours, during the transition phase where \netnarrow models are coupled to initialize a \netwide model, training is treated as an entirely new run. 
The optimizer state is not carried over, and the learning rate schedule ---including the 5-epoch linear warmup and subsequent cosine decay--- is restarted from scratch.

We consider DeiT-Base as the target model. With our \ours training protocol and one recursive step, we first train DeiT-Small models and couple them to obtain a DeiT-Base model. With two recursive steps, \ours couples DeiT-Tiny models to obtain the DeiT-Small models, further coupled into a DeiT-Base model. This whole training is therefore denoted Tiny$\rightarrow$Small$\rightarrow$Base. We compare our results with a baseline that corresponds to the training of a DeiT-Base model with the standard protocol.  To visualize the efficiency gains, we plot the test accuracy against the total training cost expressed as normalized \flops.

\begin{figure}[t!]
    \centering
        \includegraphics[width=0.7\linewidth]{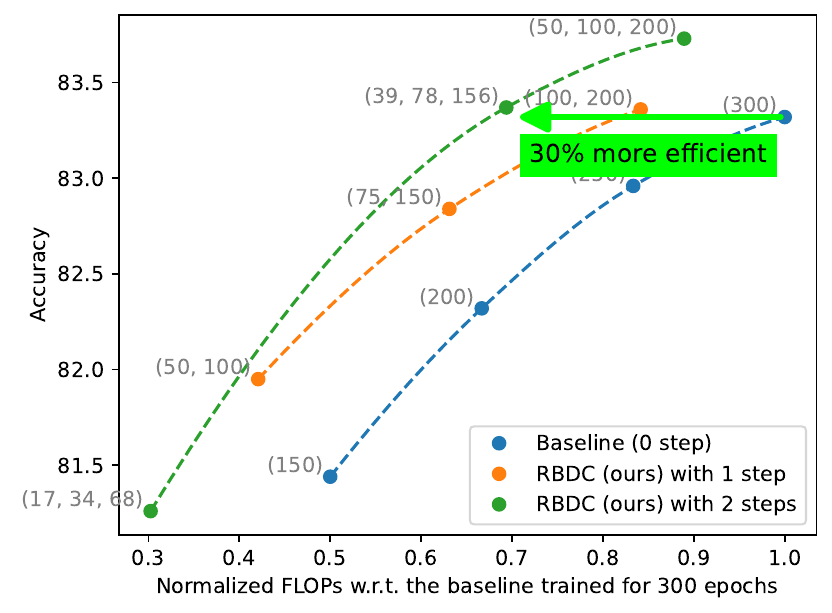}
    \caption{
        \textbf{Resource Efficiency of \ours.} 
        Accuracy \vs \flops trade-off curves when training a target DeiT-Base model. The values written in gray correspond to the number of training epochs for each recursive step, from the \netnarrowest to the target model. \ours with two recursive steps (green) is more efficient than \ours with one recursive step (orange), which is better than the standard training baseline (blue), yielding a 30\% \flops decrease for equivalent test accuracies.%
    }
    \label{fig:hierar}
\end{figure}

\textbf{Results.} The results are shown in \cref{fig:hierar}, which illustrates the trade-off curve improvements induced by \ours with one and two recursion steps over the baseline. For a given amount of recursion steps and a given number of training epochs, we obtain a point on a curve by training a DeiT-Base model with the default ratio $r=2$. The curves graphically connect the points corresponding to the same number of recursion steps but with different numbers of training epochs. Each point is thus obtained after a complete and independent run, not by further training a previous model for a few more epochs. The results demonstrate that \ours improves the efficiency over baseline training. Importantly, compared to training the baseline for 300 epochs, our results show that \ours reduces the required \flops by 30\% for a similar test accuracy (and up to 42\% at lower accuracies). Conversely, for a fixed computational budget, our recursive approach achieves higher accuracy, presumably by benefiting from the diverse features learned by the \netnarrow models.

\subsection{Training Ratio}
\label{subsec:expe-validation-design-choices}

The objective of our \second experiment is to validate the two design choices that we made in the description of \ours: the default training ratio $r=2$ between the number of epochs in successive recursive steps. The experimental setup is the same as previously.

\subsubsection{Optimal Training Ratio.}
\label{subsec:expe-validation-ratio}
\begin{figure}[t!]
    \centering
    \begin{subfigure}[b]{0.48\linewidth}
        \centering
        \includegraphics[width=\linewidth]{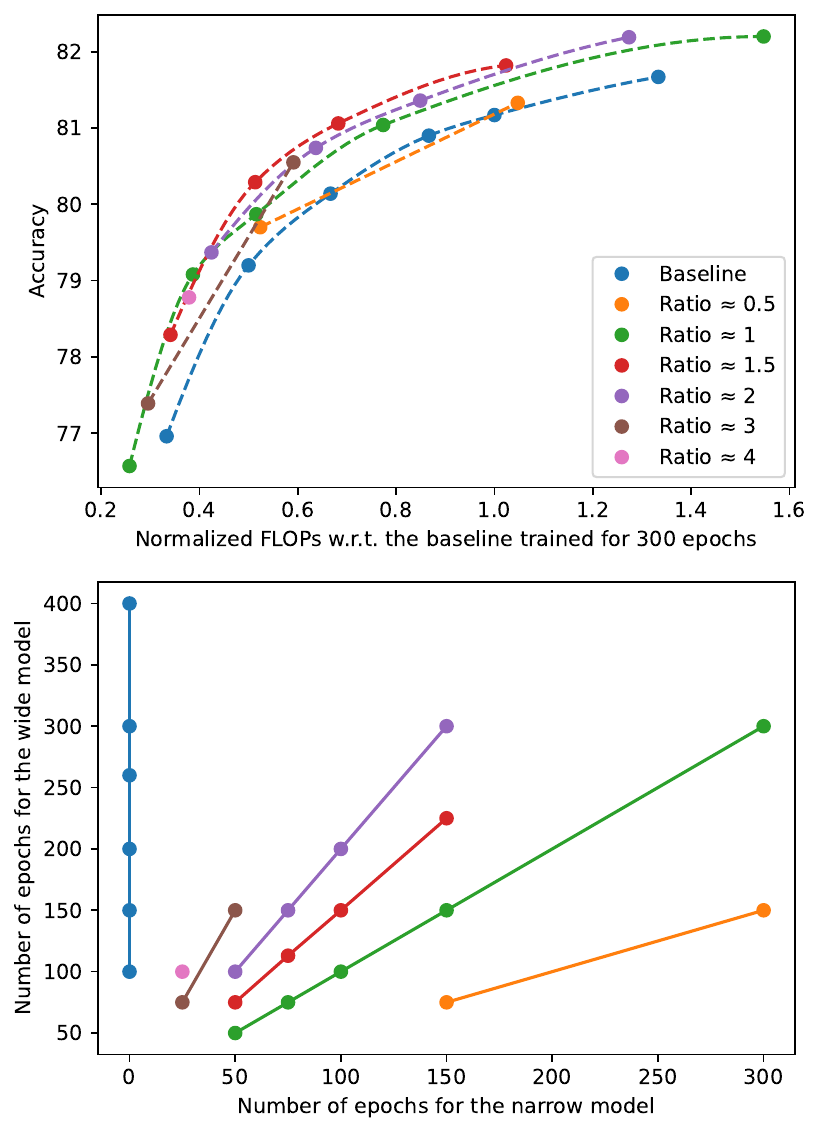}
        \caption{Impact of different training ratios with one recursive step from DeiT-Tiny to DeiT-Small.}
        \label{fig:t2s}
    \end{subfigure}
    \hfill %
    \begin{subfigure}[b]{0.48\linewidth}
        \centering
        \includegraphics[width=\linewidth]{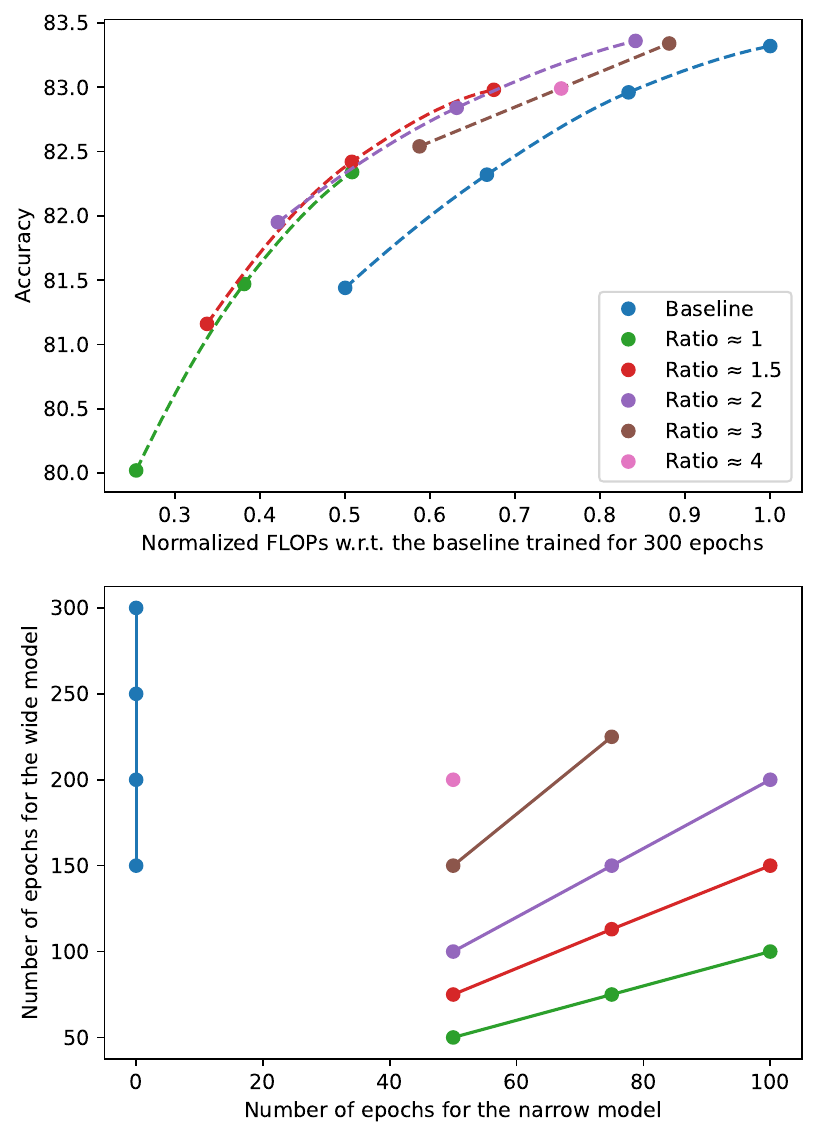}
    \caption{Impact of different training ratios with one recursive step from DeiT-Small to DeiT-Base.}
        \label{fig:s2b}
    \end{subfigure}
    
    \caption{
        \textbf{\nameref*{subsec:expe-validation-ratio}.}
        Experimental results corresponding to the training on ImageNet-1K of DeiT-Small (left) and DeiT-Base (right) models with one recursive step of \ours. 
        Top: the curves indicate an optimal ratio around the range between 1.5 and 2.0 across both model scales.
        Bottom: the repartition in training epochs of the \netnarrow and \netwide phases for each point creating lines where the slope corresponds to $r$.
    }
    \label{fig:optimal-training-ratio}
\end{figure}

Our experiments, conducted on both Tiny$\to$Small and Small$\to$Base transitions (\cref{fig:optimal-training-ratio}), reveal that the allocation of the computational budget can impact training efficiency. 
From \cref{fig:t2s,fig:s2b}, we observe that $r<1$ leads to suboptimal training, implying that the \netwide models should be trained at least as much as the \netnarrow ones. In our experiments, $r=1.5$ and $r=2$ yielded the best trade-offs between accuracy and training \flops. With $r>2$, we noticed lower gains over baseline. %

\subsection{Comparison with Network Growth Training Protocols}
\label{subsec:expe-comparison-growth}

Our $3^{rd}$ experiment benchmarks our method against the closest related literature: the network growth literature~\cite{Pham2024MixtureGrowth,Chen2015Net2Net-arxiv,Wu2020Firefly,Evci2022GradMax, Wang2023LearningToGrow, Li2026Efficient}. We divide this comparison into two parts, starting with CNNs and followed by Vision Transformers. 

\subsubsection{\underline{Comparison on CNNs.}}
Our target for this part is a ResNet-50D network, and the training is performed on ImageNet-1K.

\textbf{Implementation Details.} 
Following the protocol in~\cite{Pham2024MixtureGrowth}, we grow from a half-width ResNet-50D (0.5x) to a full-width ResNet-50D (1.0x) when using a single recursive step. For more recursive steps, we recursively apply this process, halving the initial width at each additional step (\eg, a 2-step process shrinks from 1.0x to 0.5x and then to 0.25x, and a 3-step, finally, to 0.125x), while consistently maintaining the training ratio $r=2$. Unlike MixtureGrowth~\cite{Pham2024MixtureGrowth}, which relies on template mixing and linear combinations to generate new weights, our method uses a parameter-free block-diagonal zero-initialization.

\textbf{Computational Budget and FLOPs Recalculation.} 
Importantly, the original evaluation of~\cite{Pham2024MixtureGrowth} assumes the availability of a pre-trained narrow network and excludes its computational cost from the reported efficiency metrics. 
Because \ours is designed to train target networks entirely from scratch, we recalculate the normalized FLOPs for all recombination baselines of~\cite{Pham2024MixtureGrowth}, which we report in \cref{tab:comparison}. 
\input{tables/table-cnn-comparison} 
By doing so, we account for the full training budget of every network involved. Using the fvcore library, the resulting forward costs are 8.7G FLOPs for the 1.0x target network, 2.20G FLOPs for 0.5x, 0.56G FLOPs for 0.25x, and 144.08M FLOPs for 0.125x.

The method from~\cite{Pham2024MixtureGrowth} allocates 90 epochs to the first 0.5x network, 74 epochs to the second 0.5x network, and 14 epochs to the 1.0x target network. To match this total computational budget with our default training ratio ($r=2$), we allocate our recursive training epochs as follows:
\begin{itemize}
    \item 1 recursive step: Two 0.5x networks for 22 epochs each, and the 1.0x target network for 44 epochs.
    \item 2 recursive steps: Four 0.25 networks for 10 epochs each, two 0.5x networks for 21 epochs each, and the 1.0x target network for 42 epochs.
    \item 3 recursive steps: Eight 0.125x networks for 5 epochs each, four 0.25x networks for 10 epochs each, two 0.5x networks for 20 epochs each, and the 1.0x target network for 42 epochs.
\end{itemize}

\textbf{Training Configuration.} To ensure a fair comparison with~\cite{Pham2024MixtureGrowth}, we retrain the baseline network. While~\cite{Pham2024MixtureGrowth} describes a step-decay learning rate schedule, we align the standard training protocol with the actual hyperparameters implemented in their official code. Specifically, we train a ResNet-50D using the official PyTorch Image Models (timm)~\cite{Wightman2019TIMM} pipeline, with a cosine annealing scheduler starting from a base learning rate of 0.1, preceded by a 2-epoch linear warmup, and a batch size of 512. Optimization is performed using Stochastic Gradient Descent (SGD) with a weight decay of 1e-4. As established in our DeiT experiments, the transition to a wider network triggers a full restart of the learning rate schedule and optimizer state.

\textbf{Results}. \cref{tab:comparison} shows that \ours achieves the best performance, almost matching the baseline accuracy with a FLOPs reduction of 39\%. We hypothesize that our management of the training ratio allows a flexible allocation of computational resources to the networks during training.

\subsubsection{\underline{Comparison on Vision Transformers.}}
Our target for this part is a DeiT-Base network, and the training is performed on ImageNet-1K.

\textbf{Implementation Details.} We grow the network using two recursive steps. The process starts by training DeiT-Tiny networks, which are coupled to initialize DeiT-Small networks, and these are subsequently coupled to form the final DeiT-Base target network.

\textbf{Computational Budget and FLOPs Recalculation.} To achieve our reported FLOPs reduction, we allocate the training epochs across the recursive steps while maintaining our default training ratio $r=2$ as follows: 35 epochs for the initial DeiT-Tiny networks, 70 epochs for the intermediate DeiT-Small networks, and 140 epochs for the final DeiT-Base target network. This specific allocation accounts for the total computational cost of all networks involved, which we compare against the standard 300-epoch training of the baseline DeiT-Base.

\textbf{Training Configuration.} We follow the same training protocol established in our previous DeiT experiments, with one exception: we do not use Knowledge Distillation (KD) here. This omission ensures a fair and direct comparison with the existing literature, which typically evaluates without KD.

\textbf{Results.} \cref{tab:deit_base_comparison} shows our main results. LiGO~\cite{Wang2023LearningToGrow} evaluates on DeiT-Base, achieving an 18.25\% FLOPs reduction at 81.71\% accuracy. Additionally, BTViT~\cite{xia2023budgeted} reports a 29.81\% FLOPs reduction, though it decreases the accuracy to 81.2\%. In comparison, our approach provides a 37.7\% FLOPs reduction on the same network with a slight accuracy increase (81.93\% vs. the 81.8\% baseline). We also observe a 21.8\% reduction in normalized GPU training time. 
This is lower than the FLOPs reduction mainly because our current implementation runs the networks sequentially rather than in parallel; custom kernels enabling parallel training are left to future work. 
We could not report AutoProg~\cite{Li2026Efficient} in the table because it does not provide results on DeiT-Base, nor code to produce any result.

\input{tables/table-deit-base-comparison}

\subsection{Transfer Learning on Downstream Tasks}
\label{subsec:expe-transfer-learning}

The objective of our \fourth experiment is to validate that \ours provides valuable feature representations for downstream tasks. 
We evaluate the learned DeiT-Base backbone \cite{Touvron2021Training} on both the COCO 2017 object detection and instance segmentation benchmarks~\cite{lin2014microsoft}. Throughout this section, we use the term backbone to refer specifically to the DeiT feature extractor, and model to describe the complete Mask R-CNN \cite{He2017Mask} model.

\textbf{Implementation Details.}
Our model is built upon Mask R-CNN~\cite{He2017Mask}. We integrate the pre-trained DeiT-Base backbone via the ViTDet approach~\cite{Li2022Exploring} to adapt the plain Vision Transformer (\ie, the pre-trained DeiT-Base backbone) for dense prediction tasks.  Specifically, we apply windowed attention (window size of 14) in blocks 0, 1, 3, 4, 6, 7, 9, and 10, leaving the remaining blocks as global attention to propagate information across the image.

A SimpleFPN neck \cite{Li2022Exploring} is used to build a multi-scale feature pyramid from the single-scale 768-dimensional output of the backbone. Although our \netnarrow and \netwide models, alongside the baseline, were pre-trained using the DeiT distillation objective (introducing an additional distillation token), the ViTDet architecture does not utilize the resulting distillation token. Therefore, we implemented a custom weight loader that discards this distillation token and realigns the class token with the standard patch positional embeddings during initialization.

\textbf{Training Configuration.}
The model is trained for 36 epochs using standard resizing (short edge of 800 pixels, maximum long edge of 1,333 pixels), following the baseline training recipes established for Mask R-CNN \cite{He2017Mask}. Optimization is performed using AdamW \cite{Loshchilov2019Decoupled} with a base learning rate of 1e-4, a weight decay of 0.1 and an effective batch size of 16. We apply a cosine annealing learning rate schedule \cite{Loshchilov2019Decoupled} that decays smoothly to 0.0, preceded by a linear warmup over the first 500 iterations.
To preserve the spatial features learned during pre-training, we employ layer-wise learning rate decay \cite{Bao2022BEiT, Clark2020ELECTRA} with a decay rate of 0.7 across the 12 transformer blocks of the backbone. Normalization layers, biases, and positional embeddings are excluded from the weight decay penalty.

\input{tables/table-tl-downstream}

\textbf{Results.}
We compare the downstream performance of our DeiT-Base backbone trained with \ours against the standard DeiT-Base baseline. The performance of the detection and segmentation models is measured using the \emph{mean average precision} (mAP). As detailed in \cref{tab:tl-downstream}, the model using our backbone achieves a bounding box $\text{mAP}^{\text{box}}$ of 43.4 and a mask $\text{mAP}^{\text{mask}}$ of 39.0. This outperforms the standard baseline, which yields a $\text{mAP}^{\text{box}}$ of 43.0 and a $\text{mAP}^{\text{mask}}$ of 38.5. Notably, these improvements are consistent across all intersection over union thresholds and object sizes. These results demonstrate that our efficient \ours training protocol learns useful spatial features, providing a solid initialization for downstream tasks.

%% file: tables/table-cnn-comparison.tex
\begin{table}[t]
\centering
\caption{
    \textbf{\nameref*{subsec:expe-comparison-growth}.}
    Comparison of \ours with different training protocols for CNNs on ImageNet-1K. The closest literature is about model growth methods that can be used as training from scratch protocols. The symbol $\dagger$ indicates a value reported in~\cite{Pham2024MixtureGrowth}, and $\ddagger$ stands for amounts of \flops that are higher than the values reported in~\cite{Pham2024MixtureGrowth} because, when training from scratch, we should count the amount of \flops spent to train the \netnarrow models.}
\label{tab:comparison}
\setlength{\tabcolsep}{10pt} %
\begin{tabular}{lcc}
\toprule
Protocol & Accuracy on ImageNet & Normalized \flops \\
\midrule
Standard training (baseline) & 77.79\%  & 1.00$\times$ \\
\hdashline
Net2Net \cite{Chen2015Net2Net-arxiv} & 72.29\%$^\dagger$ & 0.61$\times$$^\ddagger$ \\
Firefly \cite{Wu2020Firefly} & 71.30\%$^\dagger$ & 0.61$\times$$^\ddagger$ \\
GradMax \cite{Evci2022GradMax} & 71.73\%$^\dagger$ & 0.61$\times$$^\ddagger$ \\
MixtureGrowth \cite{Pham2024MixtureGrowth} & 74.51\%$^\dagger$ & 0.61$\times$$^\ddagger$ \\
\hdashline
\ours 1-step (Ours) & 77.32\% & 0.61$\times$ \\
\ours 2-step (Ours) & 77.41\% & 0.61$\times$ \\
\ours 3-step (Ours) & \textbf{77.55\%} & 0.61$\times$ \\
\bottomrule
\end{tabular}
\end{table}

%% file: tables/table-deit-base-comparison.tex
\begin{table}[ht]
\centering
\scriptsize
\caption{Comparison of FLOPs reduction, time reduction, and accuracy on DeiT-Base. $\dagger$ indicates a value that we have recomputed to account for the computation used for the narrower network. Baseline results are provided for fairness.}
\label{tab:deit_base_comparison}
\begin{tabular}{lccc}
\toprule
Method & FLOPs Reduction (\%) & Time Reduction (\%) & Accuracy (\%) \\
\midrule
DeiT-Base Baseline & 0.0 & 0.0 & 81.8 \\
BTViT\cite{xia2023budgeted} & 29.81 & Not provided & 81.2 \\
LiGO~\cite{Wang2023LearningToGrow} & 18.25$^\dagger$ & Not provided & 81.71 \\
\ours (Ours) & 37.7 & 21.8 & 81.93 \\
\bottomrule
\end{tabular}
\end{table}

%% file: tables/table-tl-downstream.tex
\begin{table*}[t]

\centering
\caption{
    \textbf{\nameref*{subsec:expe-transfer-learning}.}
    Object detection and instance segmentation results on the COCO 2017 validation set~\cite{lin2014microsoft} using the Mask R-CNN framework. 
    We report bounding box mean average precision ($\text{mAP}^{\text{box}}$) and mask mean average precision ($\text{mAP}^{\text{mask}}$) cross IoU thresholds and object sizes. Even though the two backbones were trained to the same classification accuracy, the one obtained with our training protocol outperforms the one obtained with the standard protocol.
}
\label{tab:tl-downstream}

\subfloat[Transfer learning: classification $\rightarrow$ detection]{
    \begin{tabular}{lcccccc}
        \toprule
        Backbone & $\text{mAP}^{\text{box}}$ & $\text{mAP}^{\text{box}}_{50}$ & $\text{mAP}^{\text{box}}_{75}$ & $\text{mAP}^{\text{box}}_{S}$ & $\text{mAP}^{\text{box}}_{M}$ & $\text{mAP}^{\text{box}}_{L}$ \\
        \midrule
        Baseline & 43.0 & 64.7 & 47.0 & 25.3 & 46.0 & 59.2 \\
        \ours (Ours) & \textbf{43.4} & \textbf{65.4} & \textbf{47.2} & \textbf{26.3} & \textbf{46.7} & \textbf{59.5} \\
        \bottomrule
    \end{tabular}
}

\subfloat[Transfer learning: classification $\rightarrow$ segmentation]{
    \begin{tabular}{lcccccc}
        \toprule
        Backbone & $\text{mAP}^{\text{mask}}$ & $\text{mAP}^{\text{mask}}_{50}$ & $\text{mAP}^{\text{mask}}_{75}$ & $\text{mAP}^{\text{mask}}_{S}$ & $\text{mAP}^{\text{mask}}_{M}$ & $\text{mAP}^{\text{mask}}_{L}$ \\
        \midrule
        Baseline & 38.5 & 61.7 & 40.7 & 18.6 & 41.4 & 59.1 \\
        \ours (Ours) & \textbf{39.0} & \textbf{62.1} & \textbf{41.1} & \textbf{19.9} & \textbf{42.0} & \textbf{59.3} \\
        \bottomrule
    \end{tabular}
}
\end{table*}

%% file: sections/5_conclusion.tex
\section{Conclusion}
\label{sec:conclusion}

We introduced \ours, an efficient training protocol to train high-capacity computer vision models from scratch. By leveraging independently trained \netnarrow models and coupling them in a block-diagonal way to serve as educated initialization for training \netwider models, \ours constructs target models recursively. That initialization is parameter-free, as it uses zero-padded interaction terms to preserve the learned features of the \netnarrow models. \ours enables to use training ratios between \netwide and \netnarrow models that  efficiently control the amount of \flops needed to train the target model.

Our results show that when fully accounting for the complete \flops budget of all training phases, our \ours training protocol improves the trade-off curve over standard training, yielding up to a 30\% \flops reduction at comparable test accuracies, and even a moderate performance improvement at similar \flops budgets. On ImageNet, we achieve competitive performance with both DeiT and ResNet models without relying on pre-trained models or parameter-heavy techniques, outperforming training protocols of the model growth literature in that regard. Furthermore, our trained models transfer successfully to downstream tasks when used as backbones, yielding improved object detection and instance segmentation performances on COCO~\cite{lin2014microsoft} over the standard baselines.

Finally, we underline that our \ours training protocol uses mechanisms that could be generalized beyond computer vision models. In particular, model architectures using mostly linear operations, such as transformers used by large language models, could also benefit from \ours. Given the cost of training such models, a training \flops reduction of 30\% could yield spectacular savings.

%% file: sections/A_0_supplementary_material.tex
\newpage
\appendix

\section{Supplementary Material}

\begingroup
\renewcommand{\\}{ }
This is the supplementary material for the paper \emph{\papertitle}.
\endgroup

\input{sections/A_1_implementation}

\input{sections/A_2_flops_epochs_relationship}

%% file: sections/A_1_implementation.tex
\subsection{Per-layer Implementation Details}

To merge two \netnarrow vision models into a \netwider model, we employ a block-diagonal initialization strategy, as motivated in~\cref{subsec:initialization}. %

Our code includes specific implementations for seven types of layers summarized in \cref{tab:layer_summary}: attention qkv projection layers, attention output projection layers, position-wise feed-forward layers, head layers, convolution layers, layer normalization layers, and batch normalization layers. 

\begin{table}[h!]
\centering
\caption{Summary of layer types and their recursive coupling initialization for each model.}
\label{tab:layer_summary}
{\setlength{\tabcolsep}{8pt}
\begin{tabular}{@{}llccc@{}}
\toprule
\textbf{\#} & \textbf{Layer Type} & \textbf{ResNet} & \textbf{DeiT} & \textbf{Coupling Eq.} \\ \midrule
1 & Attention QKV projection & & \checkmark & \cref{Eq_3}\\
2 & Attention output projection & & \checkmark & \cref{Eq_4}\\
3 & Position-wise feed-forward (MLP) & & \checkmark & \cref{Eq_5}\\
4 & Classification head & \checkmark & \checkmark & \cref{Eq_6}\\
5 & Convolution & \checkmark & & \cref{Eq_7}\\
6 & Layer Normalization & & \checkmark & \cref{Eq_8}\\
7 & Batch Normalization & \checkmark & & \cref{Eq_9}\\ \bottomrule
\end{tabular}
}
\end{table}

Our merging mechanism and initialization strategy are described hereafter: 

\begin{enumerate}
    \item \textbf{Attention QKV Projection Layers:} In standard multi-head self-attention mechanism, the query ($\mathbf{W}_Q$), key ($\mathbf{W}_K$), and value ($\mathbf{W}_V$) projections are packed into a single weight matrix. We apply the block-diagonal pattern to each projection independently. While the weight matrix $\mathbf{W}_{QKV} \in \mathbb{R}^{3d_{out} \times d_{in}}$ follows a block-diagonal structure, the bias $\mathbf{b}_{QKV} \in \mathbb{R}^{3d_{out}}$ is formed by concatenating the biases of each individual component:
    \begin{equation}
        \mathbf{W}_{QKV} = \begin{bmatrix} 
        \mathbf{W}_{Q}^{(1)} & \mathbf{0} \\ \mathbf{0} & \mathbf{W}_{Q}^{(2)} \\ \hline 
        \mathbf{W}_{K}^{(1)} & \mathbf{0} \\ \mathbf{0} & \mathbf{W}_{K}^{(2)} \\ \hline 
        \mathbf{W}_{V}^{(1)} & \mathbf{0} \\ \mathbf{0} & \mathbf{W}_{V}^{(2)} 
        \end{bmatrix}, \quad \mathbf{b}_{QKV} = \begin{bmatrix} \mathbf{b}_{Q}^{(1)} \\ \mathbf{b}_{Q}^{(2)} \\ \hline \mathbf{b}_{K}^{(1)} \\ \mathbf{b}_{K}^{(2)} \\ \hline \mathbf{b}_{V}^{(1)} \\ \mathbf{b}_{V}^{(2)} \end{bmatrix}\,. \label{Eq_3}
    \end{equation}
    
    \item \textbf{Attention Output Projection Layers:} The projection layer following the attention heads maps the concatenated features back to the embedding dimension. We %
    zero-initialize the off-diagonal interaction terms and concatenate the biases:
    \begin{equation}
        \mathbf{W}_{proj} = \begin{bmatrix} 
        \mathbf{W}_{proj}^{(1)} & \mathbf{0} \\ 
        \mathbf{0} & \mathbf{W}_{proj}^{(2)} 
        \end{bmatrix}, \quad \mathbf{b}_{proj} = \begin{bmatrix} \mathbf{b}_{proj}^{(1)} \\ \mathbf{b}_{proj}^{(2)} \end{bmatrix}\,.\label{Eq_4}
    \end{equation}

    \item \textbf{Position-wise Feed-Forward Layers:} The MLP blocks typically consist of two linear layers. Both the expansion layer ($\text{FC}_1$) and the contraction layer ($\text{FC}_2$) follow the block-diagonal structure for weights and concatenation for biases:
    \begin{equation}
        \mathbf{W}_{FC_i} = \begin{bmatrix} 
        \mathbf{W}_{FC_i}^{(1)} & \mathbf{0} \\ 
        \mathbf{0} & \mathbf{W}_{FC_i}^{(2)} 
        \end{bmatrix}, \quad \mathbf{b}_{FC_i} = \begin{bmatrix} \mathbf{b}_{FC_i}^{(1)} \\ \mathbf{b}_{FC_i}^{(2)} \end{bmatrix} \quad \text{for } i \in \{1, 2\}\,.\label{Eq_5}
    \end{equation}

    \item \textbf{Head Layers:} To aggregate the predictions from both merged branches into a single output, the final classification head is initialized as the average of the individual heads' weights and biases:
    \begin{equation}
        \mathbf{W}_{head} = \begin{bmatrix} \frac{1}{2}\mathbf{W}_{head}^{(1)} & \frac{1}{2}\mathbf{W}_{head}^{(2)} \end{bmatrix}, \quad \mathbf{b}_{head} = \frac{\mathbf{b}_{head}^{(1)} + \mathbf{b}_{head}^{(2)}}{2}\,.\label{Eq_6}
    \end{equation}

    \item \textbf{Convolution Layers:} For a weight tensor of shape ($C_{out}, C_{in}, K, K$), the spatial kernel dimensions ($K \times K$) are preserved. The merging occurs across the input and output channel dimensions, while the bias is concatenated:
    \begin{equation}
        \mathbf{W}_{conv}[:, :, k, k] = \begin{bmatrix} 
        \mathbf{W}_{conv}^{(1)}[:, :, k, k] & \mathbf{0} \\ 
        \mathbf{0} & \mathbf{W}_{conv}^{(2)}[:, :, k, k] 
        \end{bmatrix}, \quad \mathbf{b}_{conv} = \begin{bmatrix} \mathbf{b}_{conv}^{(1)} \\ \mathbf{b}_{conv}^{(2)} \end{bmatrix} \,.\label{Eq_7}
    \end{equation}

    \item \textbf{Layer Normalization:} Since layer normalization parameters are learned per-feature, we concatenate the weights ($\gamma$) and biases ($\beta$) of the two models:
    \begin{equation}
        \gamma_{wide} = \begin{bmatrix} \gamma^{(1)} \\ \gamma^{(2)} \end{bmatrix}, \quad \beta_{wide} = \begin{bmatrix} \beta^{(1)} \\ \beta^{(2)} \end{bmatrix}\,.\label{Eq_8}
    \end{equation}

    \item \textbf{Batch Normalization:} Similar to layer normalization, we concatenate the learnable affine parameters. Additionally, we concatenate the running statistics (mean and variance) to ensure the model preserves the internal activations of the narrow models:
    \begin{equation}
        \mu_{wide} = \begin{bmatrix} \mu^{(1)} \\ \mu^{(2)} \end{bmatrix}, \quad \sigma^2_{wide} = \begin{bmatrix} \sigma^{2(1)} \\ \sigma^{2(2)} \end{bmatrix}\,.\label{Eq_9}
    \end{equation}

\end{enumerate}

%% file: sections/A_2_flops_epochs_relationship.tex
\newcommand{\mytimes}{\,}

\subsection{Link between \flops budget, training ratio, and epochs}
As presented in \cref{subsec:schedule}, the amount of \flops budget is tied to the training ratio  $r$ and the number of epochs used in each stage of \ours. In this section, we show how to determine the amount of epochs necessary at each stage of \ours when the \flops budget, the number of recursive steps, and the training ratio $r$ are given.

To implement the \ours protocol under a specific computational constraint, it is necessary to determine the number of training epochs allocated to each model. The protocol operates through recursive steps of width halving, which conceptually begin with the target model and end with the \netnarrowest models, even though the training sequentially progresses from the \netnarrowest models up to the target model. 

We define the following terms to formalize the relation between the available computational budget and the training epochs:
\begin{itemize}
    \item $S$: number of width halvings.
    \item $i$: current recursive step index ($i \in \{0, \dots, S\}$), where $i=0$ corresponds to the final target model and $i=S$ corresponds to the initial \netnarrowest models.
    \item $\text{epochs}_{target}$: number of training epochs allocated to the target model.
    \item $\text{epochs}_{i}$: number of training epochs allocated to a single model at recursive step $i$.
    \item $r$: training ratio, which dictates the proportion of training epochs between a \netwide model at step $i-1$ and its constituent \netnarrow models at step $i$.
    \item $\text{FLOPs}_{total}$: computational budget to train all models in the entire pipeline.
    \item $\text{FLOPs}_{training_i}$: computational cost to fully train one model at recursive step $i$.
    \item $\text{FLOPs}_{forward_i}$: computational cost of a single forward pass for one model at recursive step $i$.
    \item $|\mathcal{D}|$: number of images in the training dataset.
\end{itemize}

The total computational cost $\text{FLOPs}_{total}$ accounts for every model involved in the pipeline. Because each model at step $i-1$ is initialized by coupling two \netnarrower models from step $i$, the total number of independent models trained at any given level $i$ is $2^i$.
The total budget is the sum of these training costs across all recursive levels:
\begin{equation}
    \text{FLOPs}_{total} = \sum_{i=0}^{S} 2^i \mytimes \text{FLOPs}_{training_i} \,.
    \label{eq:flops_all1}
\end{equation}

To estimate $\text{FLOPs}_{training_i}$, we apply the standard approximation established in the literature \cite{Hoffmann2022Training}: the backward pass requires approximately twice the compute of the forward pass to calculate gradients regarding both activations and weights. Therefore, one training step (one forward followed by one backward pass) costs three times the forward cost. This is performed $|\mathcal{D}|$ times per epoch. Hence, 

\begin{equation}
\text{FLOPs}_{training_i} = \text{epochs}_{i} \mytimes |\mathcal{D}| \mytimes 3 \mytimes \text{FLOPs}_{forward_i} \,,
\end{equation}
where $\text{FLOPs}_{forward_i}$ is obtained using the fvcore library as mentioned in \cref{sec:results}.

Given a constant training ratio $r$, the number of epochs at each step $i$ is defined relative to the target epochs $\text{epochs}_{target}$ as:
\begin{equation}
    \text{epochs}_{i} = \frac{\text{epochs}_{target}}{r^i} \,.
    \label{eq:epochs}
\end{equation}

Consequently,~\cref{eq:flops_all1} becomes

\begin{equation}
    \text{FLOPs}_{total} = \sum_{i=0}^{S} 2^i \mytimes \frac{\text{epochs}_{target}}{r^i} \mytimes |\mathcal{D}| \mytimes 3 \mytimes \text{FLOPs}_{forward_i} \,.
\end{equation}

To compute the practical training schedule for the target model given the budget $\text{FLOPs}_{total}$, we isolate $\text{epochs}_{target}$:
\begin{equation}
    \label{eq:epochs_complete}
    \text{epochs}_{target} = \frac{\text{FLOPs}_{total}}{3 \mytimes |\mathcal{D}| \mytimes \sum_{i=0}^{S} \left( \frac{2}{r} \right)^i \text{FLOPs}_{forward_i}} \,.
\end{equation}
Each training stage utilizes the full dataset (thus $|\mathcal{D}|$ images).

To generate the results from \cref{fig:hierar,fig:t2s,fig:s2b}, we calculated the amount of $\text{FLOPs}_{total}$ using a coefficient, $\alpha$, applied to the amount of \flops necessary for the baseline using a reference number of epochs, $\text{epoch}_{baseline}$. $\text{FLOPs}_{total}$ is then calculated as:
\begin{equation}
    \label{eq:alpha_flops}
    \text{FLOPs}_{total} = \alpha \mytimes \text{epoch}_{baseline} \mytimes |\mathcal{D}| \mytimes 3 \mytimes \text{FLOPs}_{forward_{0}} \,.
\end{equation}
By injecting \cref{eq:alpha_flops} in \cref{eq:epochs_complete} we have:
\begin{equation}
    \label{eq:alpha_target}
    \text{epochs}_{target} = \frac{\alpha \mytimes \text{epoch}_{baseline} \mytimes \text{FLOPs}_{forward_{0}}}{\sum_{i=0}^{S} \left( \frac{2}{r} \right)^i \text{FLOPs}_{forward_i}} \,.
\end{equation}
As models get wider and wider, the ratio of \flops between two successive forward tends to 4. This is due to linear and convolution operations being quadratic with the size of the dimension and taking bigger and bigger proportions of the amount of \flops used in the forward. To remove the necessity of computing the amount of \flops for each model, we can approximate the amount of \flops for each forward $i$ as:
\begin{equation}
    \text{FLOPs}_{forward_i} \approx \frac{\text{FLOPs}_{forward_0}}{4^i} \,.
\end{equation}
Simplifying \cref{eq:alpha_target} into:
\begin{equation}
    \text{epochs}_{target} \approx \frac{\alpha \mytimes \text{epoch}_{baseline}}{\sum_{i=0}^{S} \left( \frac{1}{2r} \right)^i} \,.
\end{equation}
Once $\text{epochs}_{target}$ is determined, the epochs for all preceding \netnarrow stages $\text{epochs}_{i}$ are derived from~\cref{eq:epochs}.

%% file: main.bbl
\begin{thebibliography}{10}
\providecommand{\url}[1]{\texttt{#1}}
\providecommand{\urlprefix}{URL }
\providecommand{\doi}[1]{https://doi.org/#1}

\bibitem{Banner2018PostT4}
Banner, R., Nahshan, Y., Soudry, D.: Post training 4-bit quantization of
  convolutional networks for rapid-deployment. In: Adv. Neural Inf. Process.
  Syst. (NeurIPS). vol.~31, pp. 7950--7958. Curran Assoc. Inc., Vancouver, Can.
  (Dec 2018)

\bibitem{Bao2022BEiT}
Bao, H., Dong, L., Piao, S., Wei, F.: {BEiT}: {BERT} pre-training of image
  transformers. In: Int. Conf. Learn. Represent. (ICLR). pp. 1--18. Virtual
  conference (Apr 2022)

\bibitem{Chen2015Net2Net-arxiv}
Chen, T., Goodfellow, I., Shlens, J.: {Net2Net}: Accelerating learning via
  knowledge transfer. arXiv  \textbf{abs/1511.05641} (2015)

\bibitem{Clark2020ELECTRA}
Clark, K., Luong, M.T., Le, Q.V., Manning, C.D.: {ELECTRA}: Pre-training text
  encoders as discriminators rather than generators. In: Int. Conf. Learn.
  Represent. (ICLR). pp. 1--18. Addis Ababa, Ethiopia (Apr 2020)

\bibitem{Evci2022GradMax}
Evci, U., van Merri{\"e}nboer, B., Unterthiner, T., Vladymyrov, M., Pedregosa,
  F.: {GradMax}: Growing neural networks using gradient information. In: Int.
  Conf. Learn. Represent. (ICLR). pp. 1--17. Virtual conference (Sept 2022)

\bibitem{facebookresearch-github-DeiT}
{Facebook Research}: Official {DeiT} repository,
  \url{https://github.com/facebookresearch/deit}

\bibitem{Gong2019Efficient}
Gong, L., He, D., Li, Z., Qin, T., Wang, L., Liu, T.: Efficient training of
  {BERT} by progressively stacking. In: Int. Conf. Mach. Learn. (ICML). Proc.
  Mach. Learn. Res., vol.~97, pp. 2337--2346. ML Res. Press (Jun 2019)

\bibitem{Gou2021Knowledge}
Gou, J., Yu, B., Maybank, S.J., Tao, D.: Knowledge distillation: A survey. Int.
  J. Comput. Vis.  \textbf{129}(6),  1789--1819 (Mar 2021)

\bibitem{Gusak2022Survey}
Gusak, J., Cherniuk, D., Shilova, A., Katrutsa, A., Bershatsky, D., Zhao, X.,
  Eyraud-Dubois, L., Shliazhko, O., Dimitrov, D., Oseledets, I., Beaumont, O.:
  Survey on efficient training of large neural networks. In: Int. Jt. Conf.
  Artif. Intell. (IJCAI). pp. 5494--5501. Vienna, Austria (Jul 2022)

\bibitem{Han2015Learning}
Han, S., Pool, J., Tran, J., Dally, W.J.: Learning both weights and connections
  for efficient neural networks. In: Adv. Neural Inf. Process. Syst. (NeurIPS).
  pp. 1135--1143. Curran Assoc. Inc., Montr{\'e}al, Can. (Dec 2015)

\bibitem{He2017Mask}
He, K., Gkioxari, G., Doll{\'a}r, P., Girshick, R.: Mask {R-CNN}. In: IEEE Int.
  Conf. Comput. Vis. (ICCV). pp. 2980--2988. IEEE, Venice, Italy (Oct 2017)

\bibitem{Henry2026LinDeps}
Henry, M., Deli{\`e}ge, A., Cioppa, A., Van~Droogenbroeck, M.: {LinDeps}: A
  fine-tuning free post-pruning method to remove layer-wise linear
  dependencies. In: Int. Conf. Pattern Recognit. (ICPR). Lect. Notes Comput.
  Sci., vol. 16816, pp. 661--677. Springer Nat. Switz., Lyon, France (Aug 2026)

\bibitem{Hoffmann2022Training}
Hoffmann, J., Borgeaud, S., Mensch, A., Buchatskaya, E., Cai, T., Rutherford,
  E., Casas, D.d.L., Hendricks, L.A., Welbl, J., Clark, A., Hennigan, T.,
  Noland, E., Millican, K., van~den Driessche, G., Damoc, B., Guy, A.,
  Osindero, S., Simonyan, K., Elsen, E., Rae, J.W., Vinyals, O., Sifre, L.:
  Training compute-optimal large language models. In: Adv. Neural Inf. Process.
  Syst. (NeurIPS). vol.~36, pp. 30016--30030. Curran Assoc. Inc., New Orleans,
  LA, USA (Nov 2022)

\bibitem{Houlsby2019Parameter}
Houlsby, N., Giurgiu, A., Jastrzebski, S., Morrone, B., De~Laroussilhe, Q.,
  Gesmundo, A., Attariyan, M., Gelly, S.: Parameter-efficient transfer learning
  for {NLP}. In: Int. Conf. Mach. Learn. (ICML). Proc. Mach. Learn. Res.,
  vol.~97, pp. 2790--2799. ML Res. Press (Jun 2019)

\bibitem{Hu2022LoRA}
Hu, E.J., Shen, Y., Wallis, P., Allen-Zhu, Z., Li, Y., Wang, S., Wang, L.,
  Chen, W.: {LoRA}: Low-rank adaptation of large language models. In: Int.
  Conf. Learn. Represent. (ICLR). pp. 1--13. Virtual conference (Sept 2022)

\bibitem{Li2026Efficient}
Li, C., Zhang, J., Lin, S., Yang, Z., Liang, J., Liang, X., Chang, X.:
  Efficient training of large vision models via advanced automated progressive
  learning. IEEE Trans. Pattern Anal. Mach. Intell. pp. 1--15 (2026)

\bibitem{Li2022Exploring}
Li, X., Duan, H., Tian, Y., Wang, F.Y.: Exploring image generation for {UAV}
  change detection. IEEE/CAA J. Autom. Sin.  \textbf{9}(6),  1061--1072 (Jun
  2022)

\bibitem{lin2014microsoft}
Lin, T.Y., Maire, M., Belongie, S., Hays, J., Perona, P., Ramanan, D.,
  Doll{\'a}r, P., Zitnick, C.L.: Microsoft {COCO}: Common objects in context.
  In: Eur. Conf. Comput. Vis. (ECCV). Lect. Notes Comput. Sci., vol.~8693, pp.
  740--755. Springer (Sept 2014)

\bibitem{loshchilov2017sgdr}
Loshchilov, I., Hutter, F.: {SGDR}: Stochastic gradient descent with warm
  restarts. In: Int. Conf. Learn. Represent. (ICLR). pp. 1--16. Toulon, France
  (Apr 2017)

\bibitem{Loshchilov2019Decoupled}
Loshchilov, I., Hutter, F.: Decoupled weight decay regularization. In: Int.
  Conf. Learn. Represent. (ICLR). pp. 1--18. New Orleans, LA, USA (May 2019)

\bibitem{Menghani2023Efficient}
Menghani, G.: Efficient deep learning: A survey on making deep learning models
  smaller, faster, and better. ACM Comput. Surv.  \textbf{55}(12),  1--37 (Mar
  2023)

\bibitem{Micikevicius2018Mixed}
Micikevicius, P., Narang, S., Alben, J., Diamos, G., Elsen, E., Garcia, D.,
  Ginsburg, B., Houston, M., Kuchaiev, O., Venkatesh, G., Wu, H.: Mixed
  precision training. In: Int. Conf. Learn. Represent. (ICLR). pp. 1--12.
  Vancouver, Can. (Apr 2018)

\bibitem{Micikevicius2022FP8Formats-arxiv}
Micikevicius, P., Stosic, D., Burgess, N., Cornea, M., Dubey, P.,
  Grisenthwaite, R., Ha, S., Heinecke, A., Judd, P., Kamalu, J., Mellempudi,
  N., Oberman, S., Shoeybi, M., Siu, M., Wu, H.: {FP8} formats for deep
  learning. arXiv  \textbf{abs/2209.05433} (2022)

\bibitem{Pham2024MixtureGrowth}
Pham, C., Teterwak, P., Nelson, S., Plummer, B.A.: {MixtureGrowth}: Growing
  neural networks by recombining learned parameters. In: IEEE/CVF Winter Conf.
  Appl. Comput. Vis. (WACV). pp. 2788--2797. IEEE, Waikoloa, HI, USA (Jan 2024)

\bibitem{Shen2024OnEfficient}
Shen, L., Sun, Y., Yu, Z., Ding, L., Tian, X., Tao, D.: On efficient training
  of large-scale deep learning models. ACM Comput. Surv.  \textbf{57}(3),
  1--36 (Nov 2024)

\bibitem{Sustainability2026Resource}
{Sustainability Directory}: Resource-efficient machine learning.
  \url{https://climate.sustainability-directory.com/term/resource-efficient-machine-learning}
  (2026)

\bibitem{Sze2017Efficient}
Sze, V., Chen, Y.H., Yang, T.J., Emer, J.S.: Efficient processing of deep
  neural networks: A tutorial and survey. Proc. IEEE  \textbf{105}(12),
  2295--2329 (Dec 2017)

\bibitem{Thompson2024AModel}
Thompson, N., Fleming, M., Tang, B.J., Pastwa, A.M., Borge, N., Goehring, B.C.,
  Das, S.: A model for estimating the economic costs of computer vision systems
  that use deep learning. In: AAAI Conf. Artif. Intell. vol.~38, pp.
  23012--23018. Assoc. Adv. Artif. Intell. (AAAI), Vancouver, Can. (Mar 2024)

\bibitem{Thompson2020TheComputational-arxiv}
Thompson, N.C., Greenewald, K., Lee, K., Manso, G.F.: The computational limits
  of deep learning. arXiv  \textbf{abs/2007.05558} (2020)

\bibitem{Touvron2021Training}
Touvron, H., Cord, M., Douze, M., Massa, F., Sablayrolles, A., J{\'e}gou, H.:
  Training data-efficient image transformers \& distillation through attention.
  In: Int. Conf. Mach. Learn. (ICML). Proc. Mach. Learn. Res., vol.~139, pp.
  10347--10357. ML Res. Press, Virtual Conf. (Jul 2021)

\bibitem{Touvron2022DeiT}
Touvron, H., Cord, M., J{\'e}gou, H.: {DeiT III}: Revenge of the {ViT}. In:
  Eur. Conf. Comput. Vis. (ECCV). Lect. Notes Comput. Sci., vol. 13684, pp.
  516--533. Springer Nat. Switz., Tel Aviv, Isra{\"e}l (2022)

\bibitem{Wang2023LearningToGrow}
Wang, P., Panda, R., Hennigen, L.T., Greengard, P., Karlinsky, L., Feris, R.,
  Cox, D.D., Wang, Z., Kim, Y.: Learning to grow pretrained models for
  efficient transformer training. In: Int. Conf. Learn. Represent. (ICLR). pp.
  1--18. Kigali, Rwanda (May 2023)

\bibitem{Wightman2019TIMM}
Wightman, R.: Pytorch image models.
  \url{https://github.com/rwightman/pytorch-image-models} (2019)

\bibitem{Wu2020Firefly}
Wu, L., Liu, B., Stone, P., Liu, Q.: Firefly neural architecture descent: a
  general approach for growing neural networks. In: Adv. Neural Inf. Process.
  Syst. (NeurIPS). vol.~33, pp. 22373--22383. Curran Assoc. Inc., Virtual
  conference (Dec 2020)

\bibitem{xia2023budgeted}
Xia, Z., Pan, X., Jin, X., He, Y., Xue', H., Song, S., Huang, G.: Budgeted
  training for vision transformer. In: Int. Conf. Learn. Represent. (ICLR). pp.
  1--16. Kigali, Rwanda (May 2023)

\bibitem{Xu2026ParameterEfficient}
Xu, L., Xie, H., Qin, S.J., Tao, X., Wang, F.L.: Parameter-efficient
  fine-tuning methods for pretrained language models: A critical review and
  assessment. IEEE Trans. Pattern Anal. Mach. Intell. pp. 1--20 (2026)

\bibitem{Zagoruyko2016Wide}
Zagoruyko, S., Komodakis, N.: Wide residual networks. In: Br. Mach. Vis. Conf.
  (BMVC). pp. 87.1--87.12. Br. Mach. Vis. Assoc. (BMVA), York, UK (Sept 2016)

\end{thebibliography}
